\documentclass[letterpaper, 10 pt, conference]{ieeeconf}  

\IEEEoverridecommandlockouts                              

\usepackage{graphicx} 

\usepackage{amsmath} 
\usepackage{amssymb}  
\usepackage{pifont} 
\usepackage{bbding}
\usepackage{xcolor}  
\definecolor{mygreen}{HTML}{0d8347}
\definecolor{myred}{HTML}{d15c3d}
\usepackage{hyperref}
\usepackage{multirow}
\usepackage{booktabs}
\usepackage{cuted}
\usepackage{float}
\usepackage{makecell}
\usepackage{subcaption}
\usepackage{tabularx}
\usepackage{caption}
\usepackage{threeparttable}
\usepackage{stfloats}
\usepackage{hyperref}
\usepackage{url}
\newcommand{\xmark}{\ding{55}}
\title{\LARGE \bf
LeHome: A Simulation Environment for \\ Deformable Object Manipulation in Household Scenarios
}

\author{Zeyi Li$^{2}$, Yushi Yang$^{1}$, Shawn Xie$^{3}$, Kyle Xu$^{3}$, Tianxing Chen$^{4}$, Yuran Wang$^{1}$, Zhenhao Shen$^{1}$, \\ Yan Shen$^{1}$, Yue Chen$^{1}$, Wenjun Li$^{4}$, Yukun Zheng$^{4}$, Chaorui Zhang$^{3}$, Siyi Lin$^{3}$, Fei Teng$^{3}$, \\ Hongjun Yang$^{2}$, Ming Chen$^{3}$, Steve Xie$^{3}$ and Ruihai Wu$^{1, *}$ %
\thanks{$^{*}$Corresponding author: Ruihai Wu, {\tt\small wuruihai@pku.edu.cn}}%
\thanks{$^{1}$Peking University.}%
\thanks{$^{2}$Institute of Automation, Chinese Academy of Sciences.}%
\thanks{$^{3}$Lightwheel.}%
\thanks{$^{4}$The University of Hong Kong.}%
}

\begin{document}
\maketitle
\thispagestyle{empty}
\pagestyle{empty}
\vspace{-10mm}
\begin{strip}
\vspace{-29mm}
    \centering
    \includegraphics[width=\textwidth]{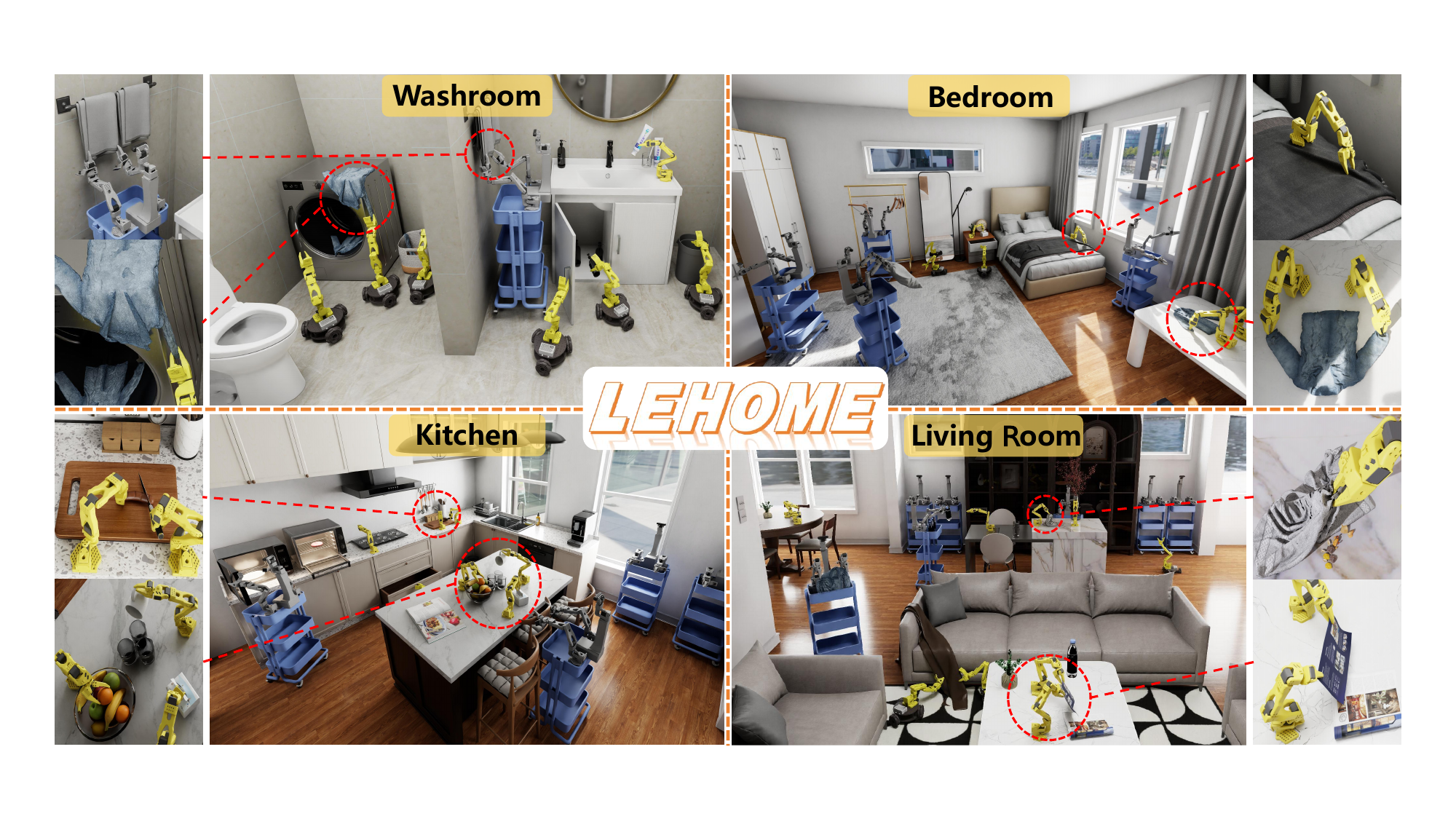}
    \vspace{-10mm}
    \captionof{figure}{\textbf{LeHome} provides a high-fidelity simulation platform by integrating various household scenarios and various objects within the scenarios, especially \textbf{deformable} objects. }  
    \label{fig:teaser}
\end{strip}
\vspace{-10mm}
\begin{abstract}

Household environments present one of the most common, impactful yet challenging application domains for robotics. Within household scenarios, manipulating deformable objects is particularly difficult, both in simulation and real-world execution, due to varied categories and shapes, complex dynamics, and diverse material properties, as well as the lack of reliable deformable-object support in existing simulations. We introduce LeHome, a comprehensive simulation environment designed for deformable object manipulation in household scenarios. LeHome covers a wide spectrum of deformable objects, such as garments and food items, offering high-fidelity dynamics and realistic interactions that existing simulators struggle to simulate accurately. Moreover, LeHome supports multiple robotic embodiments and emphasizes low-cost robots as a core focus, enabling end-to-end evaluation of household tasks on resource-constrained hardware. By bridging the gap between realistic deformable object simulation and practical robotic platforms, LeHome provides a scalable testbed for advancing household robotics. 
Webpage: \href{https://lehome-web.github.io/}{lehome-web.github.io/}.

\end{abstract}
\section{INTRODUCTION}

For humans, the household scenario is the living space with the highest frequency of daily activities and the most critical needs.
It encompasses a wide variety of essential daily tasks, such as organizing personal belongings, preparing food, and managing clothing.
Unlike those in more structured scenes such as industries, these tasks feature distinct challenges: interactions with diverse, non-standardized objects and adaptation to unstructured, dynamic environments. 

\begin{table*}[htbp]
  \centering
  \caption{\textbf{Features Comparisons} across different open-source robotics environment.}
  \label{tab1:household_framework_comparison}
  \begin{tabular}{lccccccc}
    \toprule
    Feature                          & LeHome (ours) & RoboTwin2.0 & DexGarmentLab & Behavior-1K & Libero & RLbench & Robocasa \\
    \midrule
    Household Scenarios         & \textcolor{mygreen}{\checkmark} & \textcolor{myred}{\xmark} & \textcolor{mygreen}{\checkmark} & \textcolor{mygreen}{\checkmark} & \textcolor{mygreen}{\checkmark} & \textcolor{myred}{\xmark} & \textcolor{mygreen}{\checkmark} \\
    Photorealistic Rendering         & \textcolor{mygreen}{\checkmark} & \textcolor{mygreen}{\checkmark} & \textcolor{mygreen}{\checkmark} & \textcolor{mygreen}{\checkmark} & \textcolor{myred}{\xmark} & \textcolor{myred}{\xmark}  & \textcolor{mygreen}{\checkmark} \\
    Food Deformation Simulation      & \textcolor{mygreen}{\checkmark} & \textcolor{myred}{\xmark} & \textcolor{myred}{\xmark} & \textcolor{myred}{\xmark} & \textcolor{myred}{\xmark} & \textcolor{myred}{\xmark} & \textcolor{myred}{\xmark} \\
    Flame and Particle Simulation    & \textcolor{mygreen}{\checkmark} & \textcolor{myred}{\xmark} & \textcolor{myred}{\xmark} & \textcolor{mygreen}{\checkmark} & \textcolor{myred}{\xmark} & \textcolor{myred}{\xmark} & \textcolor{myred}{\xmark} \\
    Fluid Simulation                 & \textcolor{mygreen}{\checkmark} & \textcolor{myred}{\xmark} & \textcolor{mygreen}{\checkmark} & \textcolor{mygreen}{\checkmark} & \textcolor{myred}{\xmark} & \textcolor{myred}{\xmark} & \textcolor{myred}{\xmark} \\
    Garment Manipulation               & \textcolor{mygreen}{\checkmark} & \textcolor{myred}{\xmark} & \textcolor{mygreen}{\checkmark} & \textcolor{mygreen}{\checkmark} & \textcolor{myred}{\xmark} & \textcolor{myred}{\xmark} & \textcolor{myred}{\xmark} \\
    Articulation and Rigid Manipulation& \textcolor{mygreen}{\checkmark} & \textcolor{mygreen}{\checkmark} & \textcolor{myred}{\xmark} & \textcolor{mygreen}{\checkmark} & \textcolor{mygreen}{\checkmark} & \textcolor{mygreen}{\checkmark} & \textcolor{mygreen}{\checkmark} \\
    Multi-material Manipulation      & \textcolor{mygreen}{\checkmark} & \textcolor{myred}{\xmark} & \textcolor{mygreen}{\checkmark}& \textcolor{mygreen}{\checkmark} & \textcolor{myred}{\xmark} & \textcolor{myred}{\xmark} & \textcolor{myred}{\xmark} \\
    Teleoperation                    & \textcolor{mygreen}{\checkmark} & \textcolor{myred}{\xmark} & \textcolor{myred}{\xmark} & \textcolor{mygreen}{\checkmark} & \textcolor{myred}{\xmark} & \textcolor{myred}{\xmark} & \textcolor{mygreen}{\checkmark} \\
    Low-cost Robot                   & \textcolor{mygreen}{\checkmark} & \textcolor{myred}{\xmark} & \textcolor{myred}{\xmark} & \textcolor{myred}{\xmark} & \textcolor{myred}{\xmark} & \textcolor{myred}{\xmark} & \textcolor{myred}{\xmark} \\
    \bottomrule
  \end{tabular}
\end{table*}

Amid these challenges, household scenarios have attracted increasing attention, ranging from the development of simulated household simulation platforms~\cite{li2021igibson, li2023behavior, szot2021habitat} to the design of robotic policies for household tasks~\cite{bjorck2025gr00t,jiang2025behavior,yenamandra2023homerobot}. 

However, a key mismatch remains between current benchmarks and real-world needs: most existing methods target rigid and articulated objects, while many household tasks inherently involve diverse deformable objects (\emph{e.g.}, garments and food), for which current support remains limited. Such objects lack fixed shapes, deform nonlinearly under applied forces, and exhibit dynamic physical parameters that are difficult to characterize. Consequently, creating large-scale, realistic, and diverse deformable-object datasets for training robust policies remains a major challenge, due to two fundamental issues: (i) collecting real-world household data is prohibitively expensive and labor-intensive, as deformable objects’ variable states and the inherent complexity of household environments make it difficult to obtain sufficient high-quality data; and (ii) achieving accurate and authentic modeling of deformation is intrinsically hard, since it requires simultaneously capturing complex material properties, nonlinear dynamics, and realistic interactions.

As shown in Fig.~\ref{fig:teaser}, we present \textbf{LeHome}, a household simulation environment designed for high-fidelity manipulation of diverse deformable objects. LeHome provides a household asset library spanning liquids, gaseous fluids, granular objects, linear objects, thin shells, and volumetric objects. To model these categories, LeHome combines appropriate simulation engines, including Position-Based Dynamics (PBD), Finite Element Method (FEM), and Eulerian Fluid Simulation, to improve physical realism while preserving broad task coverage. Furthermore, to enable complex manipulation mechanisms, such as a sausage being cut into multiple pieces with a knife. LeHome introduces a graphical logical modeling method called the \textbf{Action Graph}, representing causal mechanisms. This method enforces causal consistency between actions and outcomes, ensuring the simulation aligns with real-world dynamics. In summary, these designs allow LeHome to provide a reliable physical foundation for deformable object manipulation and establish a strong basis for generating high-quality training data.

In terms of supported robot platforms, LeHome spans multiple robotic embodiments, ranging from mainstream commercial manipulators (e.g., UR, Franka) to open-source low-cost platforms (e.g., LeRobot, XLeRobot). Compared with widely-used industrial robots, low-cost robots are more compact, easier to deploy, and cheaper to maintain. These properties make them better suited for scalable deployment in household environments. Therefore, LeHome places particular emphasis on supporting a diverse set of open-source, low-cost robot arms, which are easy to set up, modify, and maintain, while still providing sufficient manipulation capability for everyday household tasks. By lowering the hardware and deployment barriers, this emphasis enables scalable, repeatable validation of core household-robot functionalities in realistic home environments, and in turn supports broader adoption and large-scale deployment.

Finally, to further strengthen LeHome in both scalable data acquisition and user-friendliness, we integrate diverse teleoperation methods, offering multiple devices to fit different demands and allowing anyone to operate it intuitively. Moreover, it is compatible with both virtual and real scenarios, enabling simulation datasets to be seamlessly supplemented with real-world demonstrations under the same pipeline, which in turn supports efficient collection of manually annotated data. Meanwhile, we employ domain randomization to augment existing demonstrations by injecting randomized scene variations during trajectory replay, thereby automatically generating more diverse data with richer visual appearances and interaction conditions. Real-robot evaluations validate that LeHome supports effective sim-to-real transfer for diverse deformable object manipulation tasks, and that domain randomization mitigates distribution shift, resulting in more robust real-world deployment.
\section{RELATED WORK}
\begin{figure*}[htbp]
    \centering
    \includegraphics[width=\textwidth]{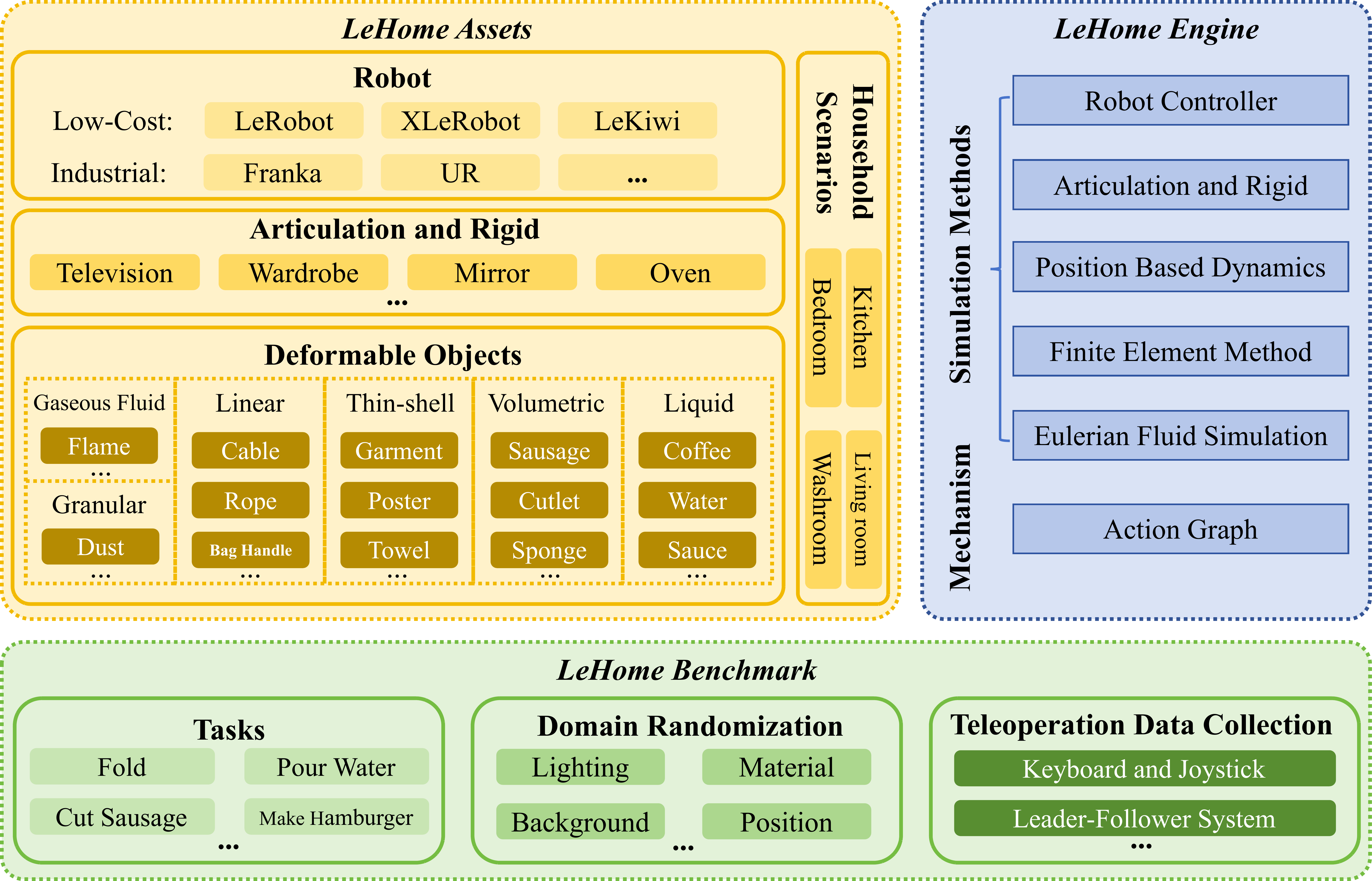}
    \caption{\textbf{The Architecture of LeHome}. (Left) LeHome Assets can deliver realistic simulations of different robots, articulation/rigid objects, deformable objects, and cover multiple household scenarios. (Right) Leveraging diverse simulation methods and mechanisms, LeHome Engine enables various simulation capabilities. (Bottom) Subsequently, LeHome Benchmark utilizes these assets to construct tasks, conduct domain randomization, and facilitate teleoperation data collection.}
    \label{fig2：system}
\end{figure*}

\subsection{Simulation and Manipulation in Household Scenarios}
Household scenarios are particularly valuable for robotic simulation because they closely reflect the complexities of real-world settings. Existing simulation platforms have established a solid foundation for related research: many~\cite{chen2025robotwin,geng2025roboverse,Mu_2025_CVPR,liu2023libero,james2020rlbench} show practical value in building desktop-level or single-object scenes, effectively supporting targeted research.
Some platforms~\cite{szot2021habitat,li2021igibson} further extend scene coverage to more diverse scenarios, yet their visual fidelity remains limited, falling short of near-photorealistic realism.
Additionally, studies~\cite{li2023behavior,tao2024maniskill3,nasiriany2024robocasa,gong2023arnold} have actively addressed these limitations, however, high-fidelity interactions with deformable objects (\emph{e.g.}, garments, food) remain challenging. These deformable objects are ubiquitous in daily life, and accurately simulating their manipulation is critical to realistic household scenarios.
To address these existing gaps, we introduce the LeHome platform, which integrates both flexibility and scalability. As Tab.~\ref{tab1:household_framework_comparison}, Lehome incorporates high-fidelity assets across various room types and enables complex interactions with deformable objects (\emph{e.g.}, folding garments and cutting sausages), thereby effectively addressing a key limitation in current simulation platforms.

\subsection{Deformable Object Simulation and Manipulation}
Robotic manipulation of deformable objects has recently gained attention, with research on distinct object types.

Some works~\cite{zhou2023clothesnet,lu2024garmentlab,Xue2023UniFolding,ha2022flingbot,wang2025dexgarmentlab,lin2021softgymbenchmarkingdeepreinforcement,tie2025etseed} specialize in the simulation and manipulation of cloth-related tasks. For soft materials, fluids,, and food, dedicated efforts~\cite{li2022contact,xian2023fluidlab,huang2021plasticinelab,heiden2021disect,lin2022diffskill,shi2023robocook} have also focused on their simulation and manipulation. Additionally, studies have addressed the manipulation of linear and thin-shell objects, supported by dedicated platforms~\cite{wang2023thin,seita2023learningrearrangedeformablecables}.
However, most simulators remain confined to a single category of deformable objects and cannot capture multi-modal interactions, \emph{e.g}., cloth with furniture or sausages with knives. In contrast, LeHome unifies high-quality deformable assets (e.g., garments, food and fluids), within furnished household scenarios, enabling systematic study of multi-modal deformable manipulation in realistic environments.

\subsection{Manipulation with Low-Cost Robot}

Mainstream industrial robots such as Franka and UR demonstrate strong reliability while remaining costly and bulky, with potential  obstacles to broad household deployment. A variety of low-cost and open-source platforms (\emph{e.g.}, Aloha~\cite{zhao2023learning}, Mobile Aloha~\cite{fu2024mobile}, OpenArm~\cite{OpenArm}) have emerged, providing modular robotic arms and mobile systems for household research. Among these platforms, the LeRobot series (LeRobot~\cite{cadene2024lerobot}, Lekiwi~\cite{Lekiwi}, XLeRobot~\cite{wang2025xlerobot}) has further reduced costs by integrating insights from preceding products. Moreover, supported by an open-source community, it integrates the implementation of many manipulation algorithms for deployment. Consequently, LeHome chooses to integrate this LeRobot series across diverse household scenarios, thereby offering a potential path for future large-scale deployment of household robots.
\section{ENVIRONMENT}
\begin{figure*}[htbp]
    \centering
    \includegraphics[width=\textwidth]{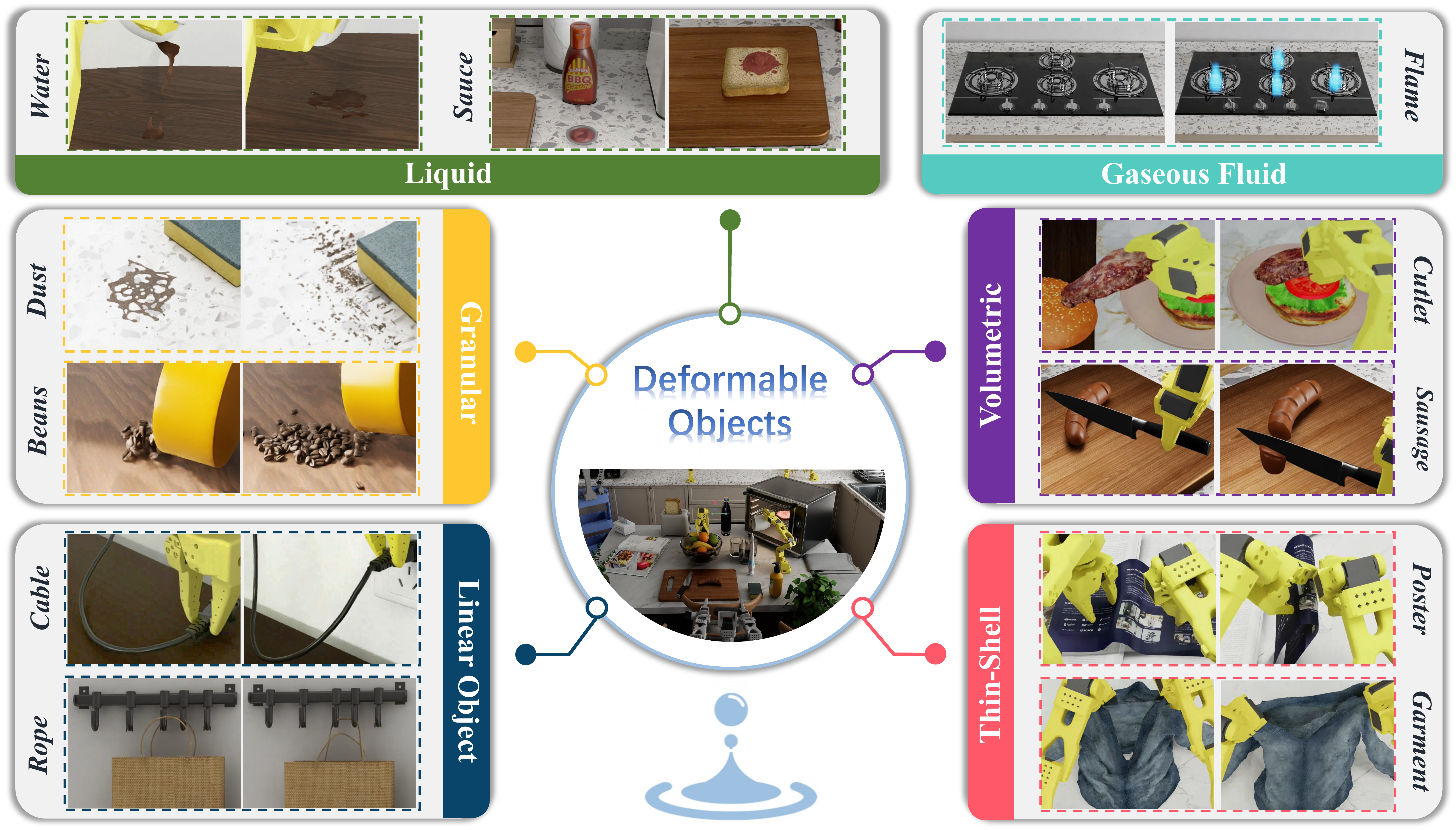}
    \caption{\textbf{Simulated Deformble Objects} cover 6 categories with visually and physically high-fidelity assets for each category.}
    \label{fig3：Assets}
\end{figure*}
\subsection{Overview}
LeHome is a household simulation environment designed for manipulation research, with a particular emphasis on deformable objects. As illustrated in Fig.~\ref{fig2：system}, \textbf{LeHome} is organized into three tightly connected components. The first is \textbf{LeHome Assets}, which provides realistic household scenes, diverse object instances, and multiple robot embodiments as the content foundation of the environment. The second is \textbf{LeHome Engine}, which serves as the computational core by integrating physics simulation and interaction mechanisms to model fine-grained manipulation dynamics. The third is \textbf{LeHome Benchmark}, which defines representative household tasks and supports teleoperation data collection for training and evaluation. Together, these components form a unified framework that combines realistic scene modeling, controllable simulation, and efficient data acquisition.

\subsection{Assets and Physics}
Beyond rigid and articulated objects, LeHome provides diverse deformable assets for household tasks. 
To model heterogeneous physical behavior in a structured manner, we categorize deformable objects into six classes according to their mechanical characteristics, as shown in Fig.~\ref{fig3：Assets}:

\begin{enumerate}

    \item \textbf{Liquid:}
    Materials that flow readily and thus have no fixed shape. While their volume is approximately conserved, they deform under shear and adapt to container shape. Examples: water, juice.

    \item \textbf{Gaseous Fluid:}
    Here, fluid specifically refers to the gaseous phase, rather than liquid. It is compressible and has no fixed volume, with density varying with time and pressure. Example: flame.

    \item \textbf{Granular Object:} 
    Composed of discrete granular aggregates, with mechanical behaviors dominated by inter-particle collision, friction, and stacking. Example: Dust, Beans.

    \item \textbf{Linear Object:} 
    Characterized by a highly slender geometry. Under outer loads, deformation is dominated by bending. Examples: cable and rope.
    \\
    \item \textbf{Thin Shell:}
    Characterized by thin planar geometry, deformation is dominated by shell bending and membrane stretching. Example: poster, garment, bag.

    \item \textbf{Volumetric Object:} 
     Governed by continuum mechanics, these are treated as continuous solids with stress/strain fields. Examples: cutlet, sausage, burger.

\end{enumerate}

To achieve both realism and efficiency, we assign simulation methods to each category based on physical characteristics and task requirements:

\begin{enumerate}

    \item \textbf{Liquid:} 
    All Liquids are simulated using PBD, which efficiently captures liquid motion and interactions under frequent contact. 
    \item \textbf{Gaseous Fluid:} For flame simulation, we leverage Omniverse Flow with a sparse voxel grid representation to update key fields, enabling efficient Eulerian Fluid Simulation with realistic visual appearance.

    \item \textbf{Granular Object:} 
Fine granules (e.g., dust) are modeled with PBD for dispersed motion, while coarse objects (e.g., coffee beans) are approximated as rigid bodies for stable contact interactions.

    \item \textbf{Linear Object:} 
We support two simulation modes for linear objects: (1) a multi-rigid-body chain model, simple and efficient; and (2) an FEM-based deformable model capturing detailed bending and stretching at higher computational cost.
   \item \textbf{Thin Shell:}
For thin-shell objects with wrinkling (e.g., garments), we use PBD for efficient simulation of stretching and bending. When wrinkling is less critical (e.g., posters), FEM is used for stable elastic responses.
    \item \textbf{Volumetric Object:} 
    All volumetric deformable objects (e.g., patty, sausage) are simulated using FEM with volumetric discretization, enabling elastic or elastoplastic stress-strain modeling.

\end{enumerate}

\begin{figure*}[htbp]
    \centering
\includegraphics[width=\textwidth]{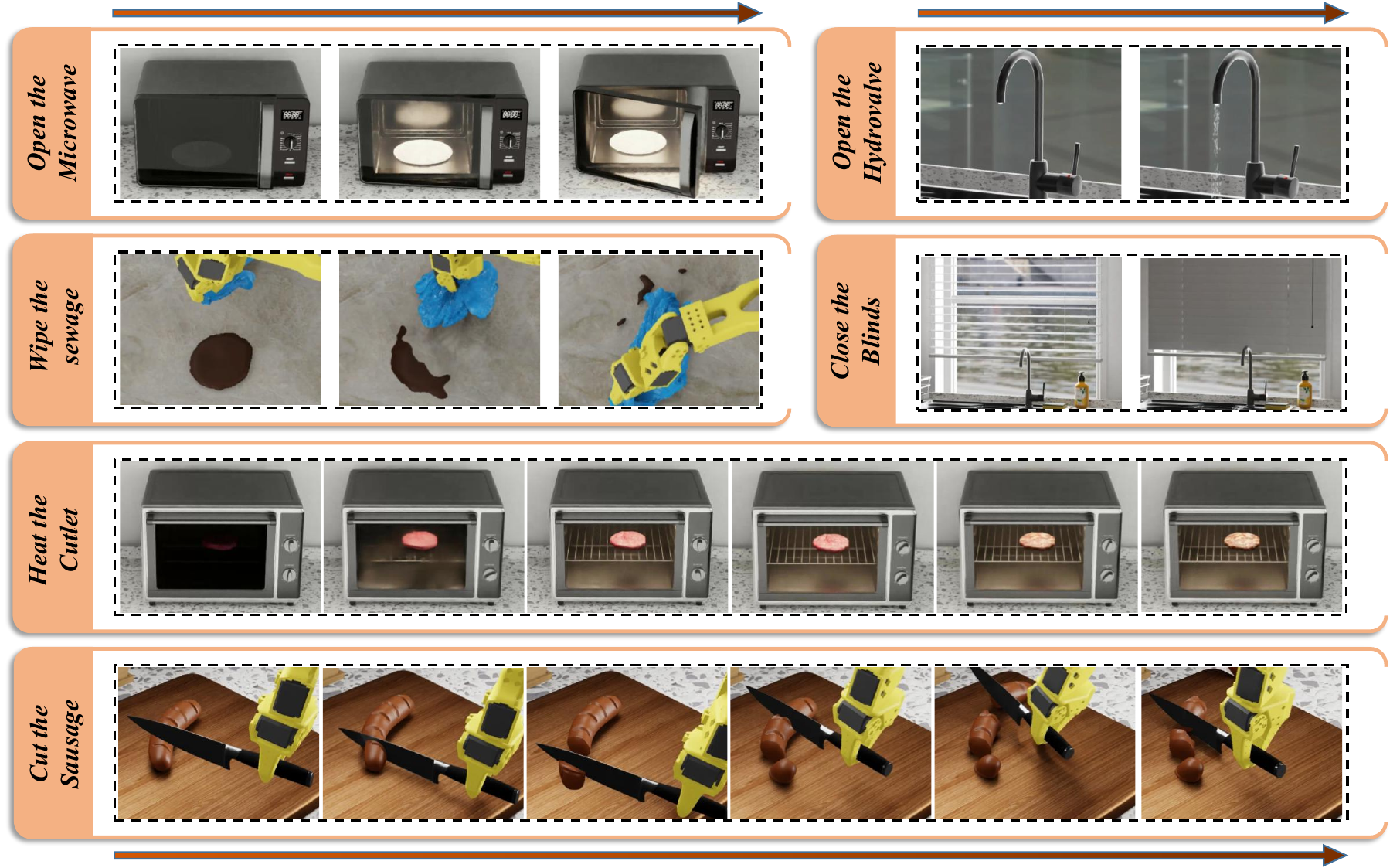}
    \caption{\textbf{Diverse Manipulation Mechanisms.} LeHome models causal relationships of manipulation through the action graph, ensuring the simulation results align with real-world causal relationships and providing high-fidelity interactions.}
    \label{fig4：mechanism}
\end{figure*}

\subsection{Mechanisms}
Previous simulation systems struggle to capture mechanisms induced by physical interactions (e.g., object morphological splitting and state transitions), and suffer from weak coupling between action logic and physical changes, making it hard to replicate real-world cause–effect dynamics. Our LeHome achieves high-fidelity dynamic mechanisms, as shown in Fig.~\ref{fig4：mechanism}, by proposing the \textbf{Action Graph}.

Specifically, the core lies in decomposing key simulation links (action triggering, state transition, and physical computation) into controllable modular units, enabling physics-compliant interactive simulation while balancing efficiency and real-time response. Supported by an ``event-response" model, Action Graph consists of three key components: 
\begin{enumerate}
    \item \textbf{Attributes:} carriers of a name, a data type, or a set of metadata (e.g., a constant or a coordinate).
    \item \textbf{Nodes:} basic units of the action graph. There are many types based on function, such as on-trigger, computation, and state update nodes.
    \item \textbf{Connections:} edges in the action graph connecting different nodes, representing the flow of data and logical relationships.
\end{enumerate}

\begin{figure}[htbp]
    \centering
    \includegraphics[width=0.4\textwidth]{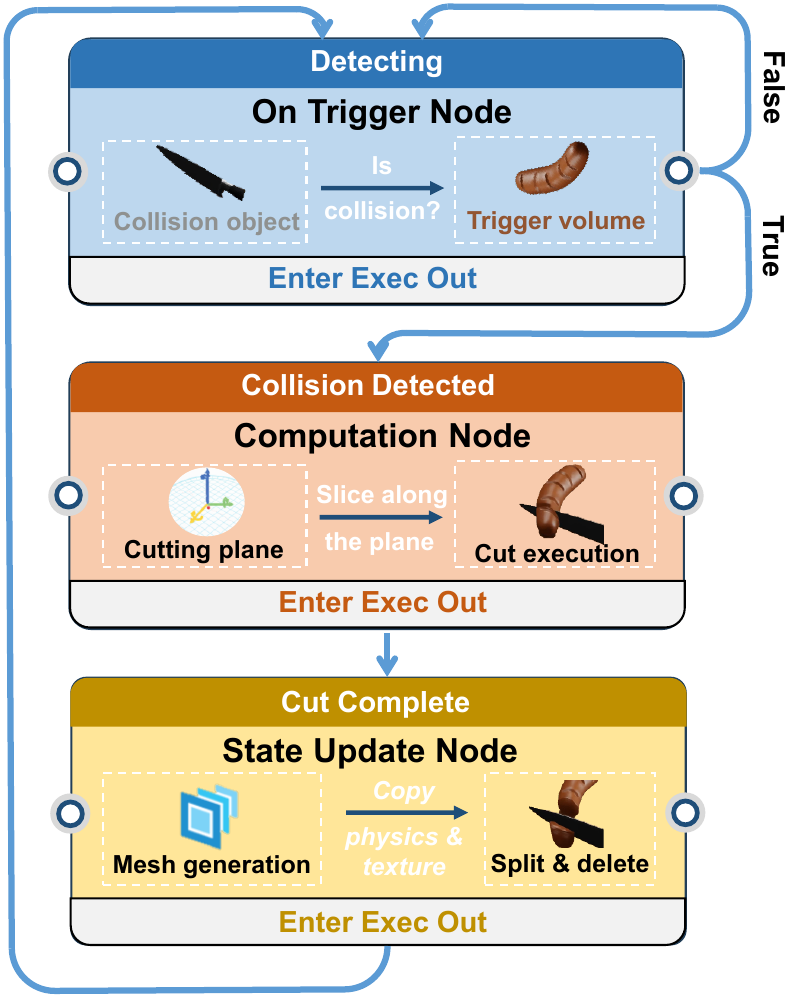}
    \caption{\textbf{Action Graph Workflow} on  sausage cutting, consisting of an On-Trigger Node, a Computation Node, and a State Update Node, connected in an ordered way.}
    \label{fig5：action graph}
    \vspace{-5mm}
\end{figure}

As shown in Fig.~\ref{fig5：action graph}, the Action Graph achieves simulation of sausage cutting as follows: First, the On Trigger Node captures collision signals between the cutting tool and the sausage. When the cutting tool collides with the sausage, a ``cut-trigger'' signal is generated. Then, the Computation Node initiates geometric mesh segmentation for the sausage model based on the cutting plane. Finally, after the mesh segmentation by the cutting computation, the State Update Node performs new object creation and updates the physical properties as well as textures of the split halves, realizing the ``Split'' process.

Compared with previous works, Action Graph provides modularity (node-level reuse), controllability (explicit dependencies), and extensibility (new mechanisms can be added by composing nodes). It is designed to support diverse household interaction patterns while remaining compatible with real-time simulation loops. 

\subsection{Robots and Teleoperation}
We provide a suite of low-cost LeRobot-family embodiments (LeRobot, LeKiwi and XLeRobot), covering single, bimanual, and mobile-manipulation configurations for household tasks.

\begin{figure}[htbp]
    \centering
    \includegraphics[width=0.48\textwidth]{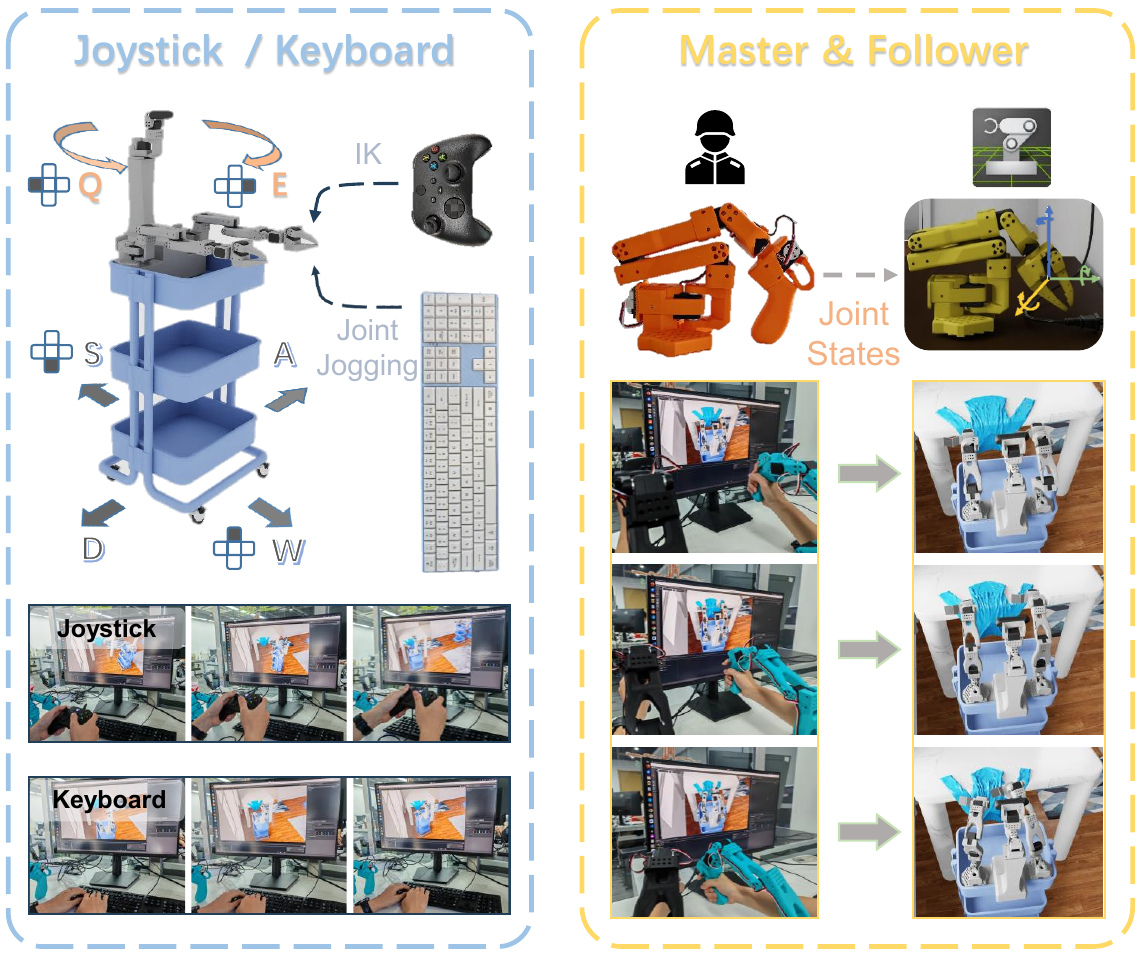}
    \caption{\textbf{Teleoperation Methods.} (Left) We use Joystick and Keyboard to teleoperate XLeRobot, and (Right) Leader-Follower Teleoperation for LeRobot.}
    \label{fig6:tele}
    \vspace{-3mm}
\end{figure}

As shown in Fig.~\ref{fig6:tele}, we provide multiple teleoperation methods for demonstration collection. We support keyboard and joystick inputs that issue joint-space commands, offering a low-cost and widely accessible control modality. We further integrate a leader-follower system that enables intuitive joint-synchronized teleoperation. For mobile robots, the system supports hybrid control by combining leader-follower arm synchronization with keyboard/joystick commands for base locomotion.

Overall, by emphasizing low-cost, compact form factor, and ease of operation for household settings, this robot lineup lowers the deployment barrier and facilitates the broader adoption of home-robot applications in everyday families.

\subsection{Domain Randomization}
To mitigate the sim-to-real gap in diverse household settings, LeHome applies domain randomization (DR) during teleoperation and further augments the dataset via trajectory replay. At the start of each episode, four factors are randomized to generate varied yet feasible configurations:
(1) \textbf{Initialization position}: objects sampled within workspace bounds while avoiding invalid states;
(2) \textbf{Lighting}: illumination intensity and color temperature vary;
(3) \textbf{Background texture}: tabletop textures sampled from a curated set; and
(4) \textbf{Material property}: visual materials (e.g., cloth/liquid appearance) randomized while preserving physical dynamics.

After data collection, we replay recorded trajectories under randomized appearance factors to further increase visual diversity. Object positions remain fixed to preserve task geometry and contacts. Trajectories are then filtered using task-specific success detectors (e.g., state validation or geometric constraints), retaining only successful ones for training. This process improves robustness with minimal additional demonstration cost.

\section{EXPERIMENT}

With the proposed LeHome environment, we conduct experiments to demonstrate that LeHome offers high-fidelity simulation (especially for deformable objects) to support policy learning across diverse household tasks, and that training in LeHome facilitates effective sim-to-real transfer and improves real-world performance.

\subsection{Tasks and Settings}

To demonstrate the capability to support different policy learning algorithms on diverse daily household tasks, we selected six representative tasks from LeHome, covering single-arm and bimanual manipulation, tool use, deformable food interaction, fluids, and rigid object manipulation, rigid-deformable and deformable-deformable interactions. The tasks span multiple rooms (bedroom, kitchen, living room, bathroom) to reflect realistic household diversity.

\begin{figure}[h]
    \centering
    \captionsetup[sub]{font=small,justification=centering}

    \begin{subfigure}[t]{0.32\linewidth}
        \centering
        \includegraphics[width=\linewidth, keepaspectratio]{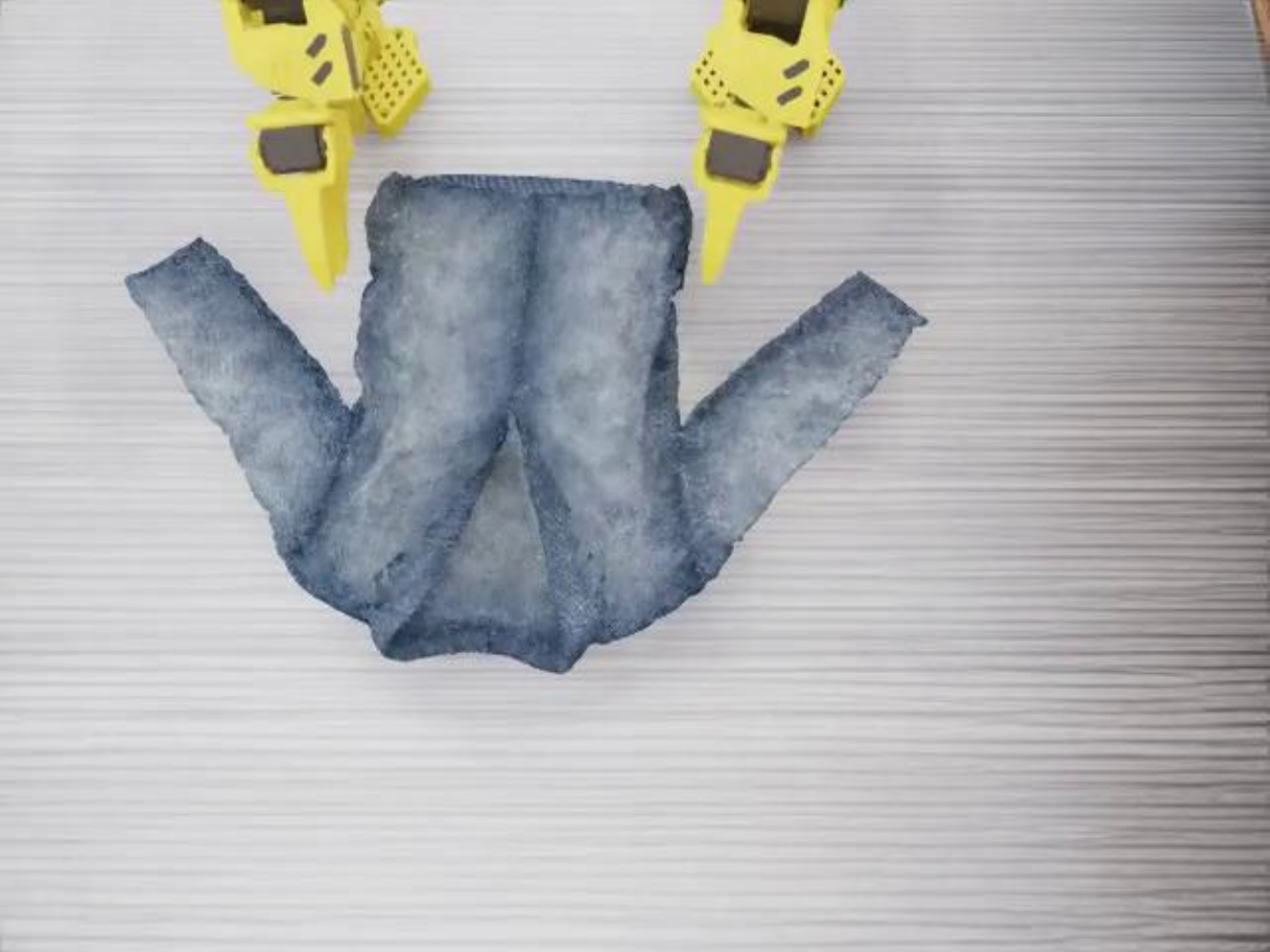}
        \caption{Fold Garment}
        \label{fig:fold}
    \end{subfigure}
    \begin{subfigure}[t]{0.32\linewidth}
        \centering
        \includegraphics[width=\linewidth, keepaspectratio]{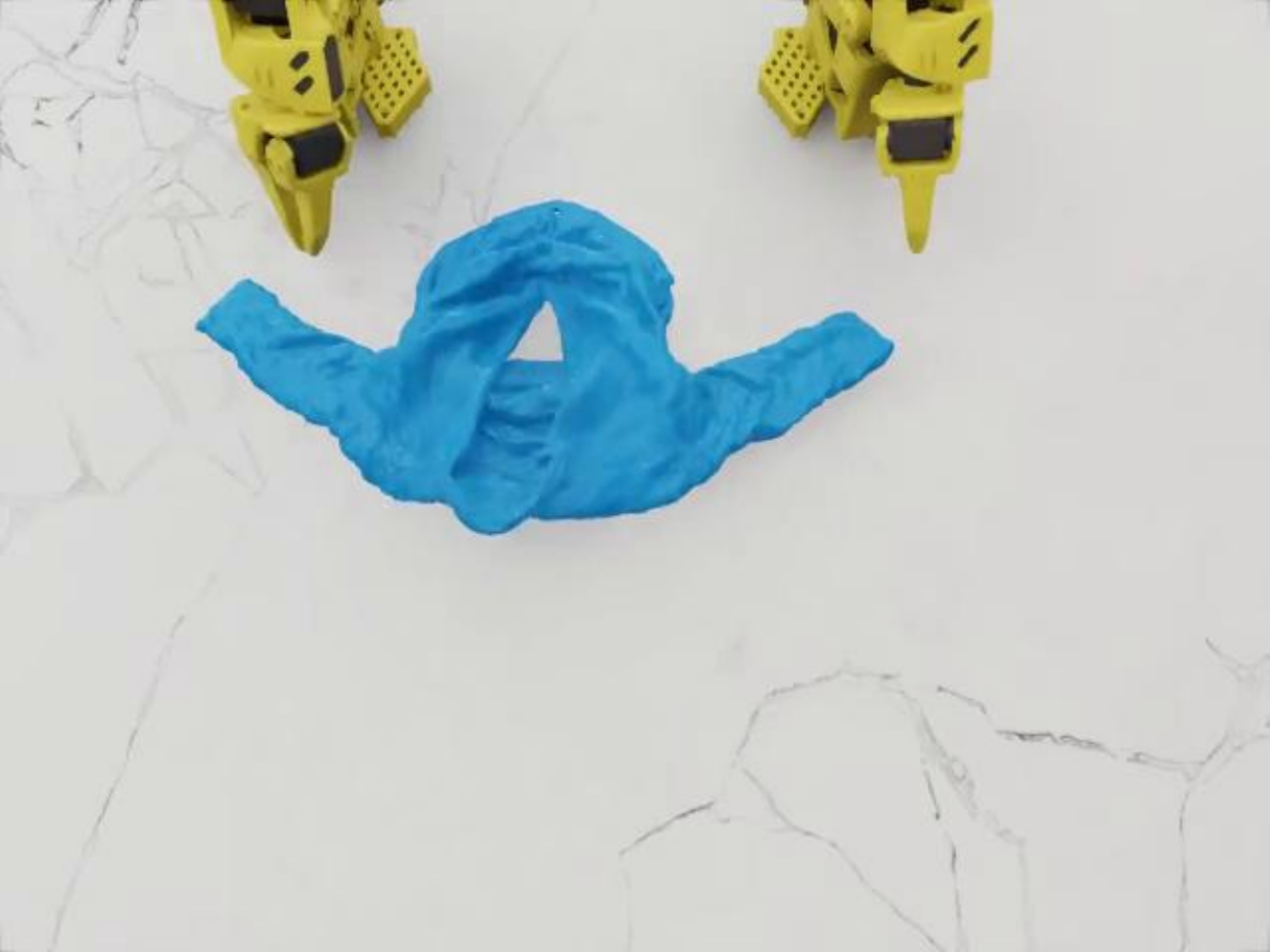}
        \caption{Fling Garment}
        \label{fig:fling}
    \end{subfigure}
    \begin{subfigure}[t]{0.32\linewidth}
        \centering
        \includegraphics[width=\linewidth, keepaspectratio]{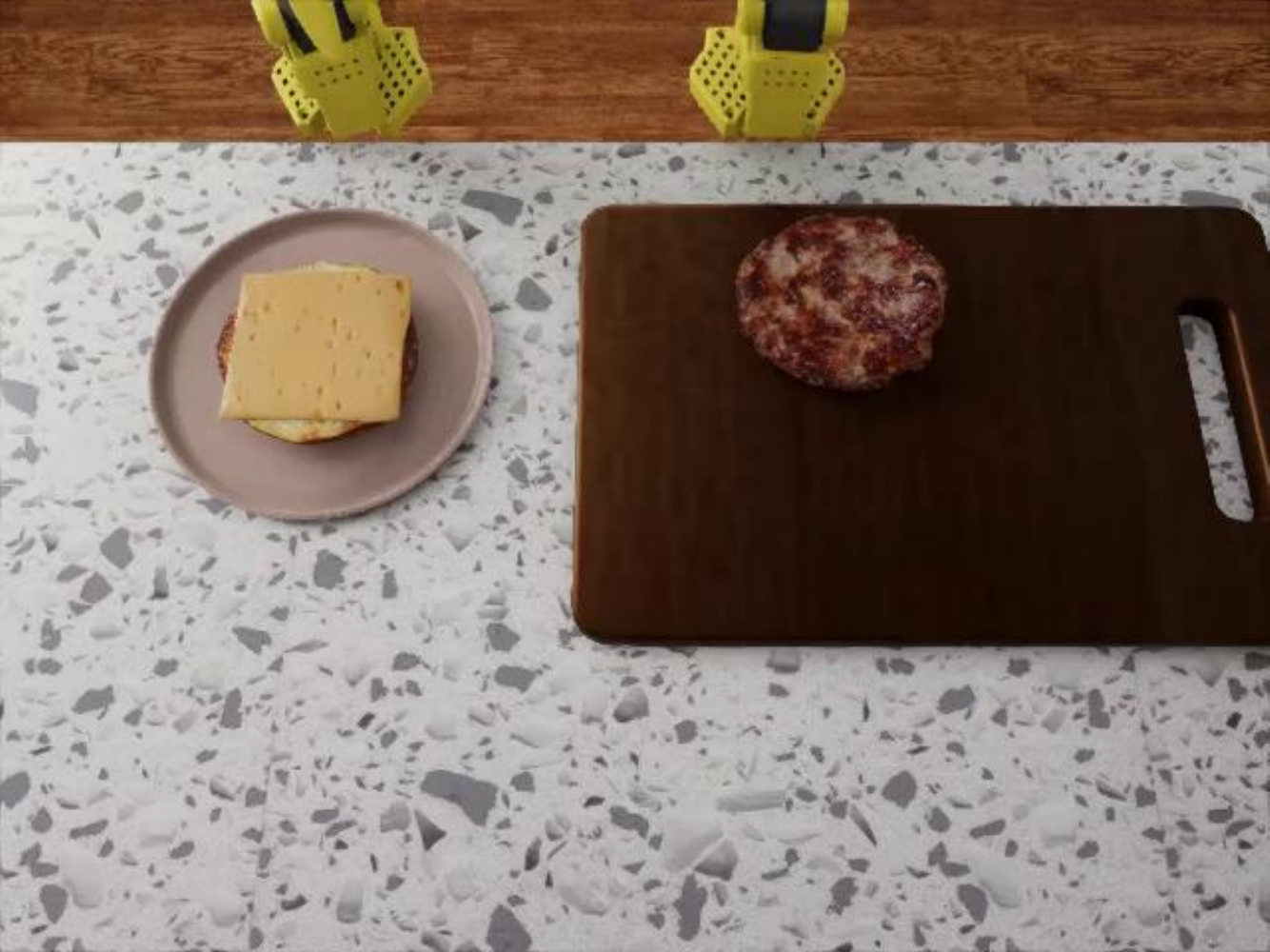}
        \caption{Assemble Burger}
        \label{fig:burger}
    \end{subfigure}

    \vspace{4pt}

    \begin{subfigure}[t]{0.32\linewidth}
        \centering
        \includegraphics[width=\linewidth, keepaspectratio]{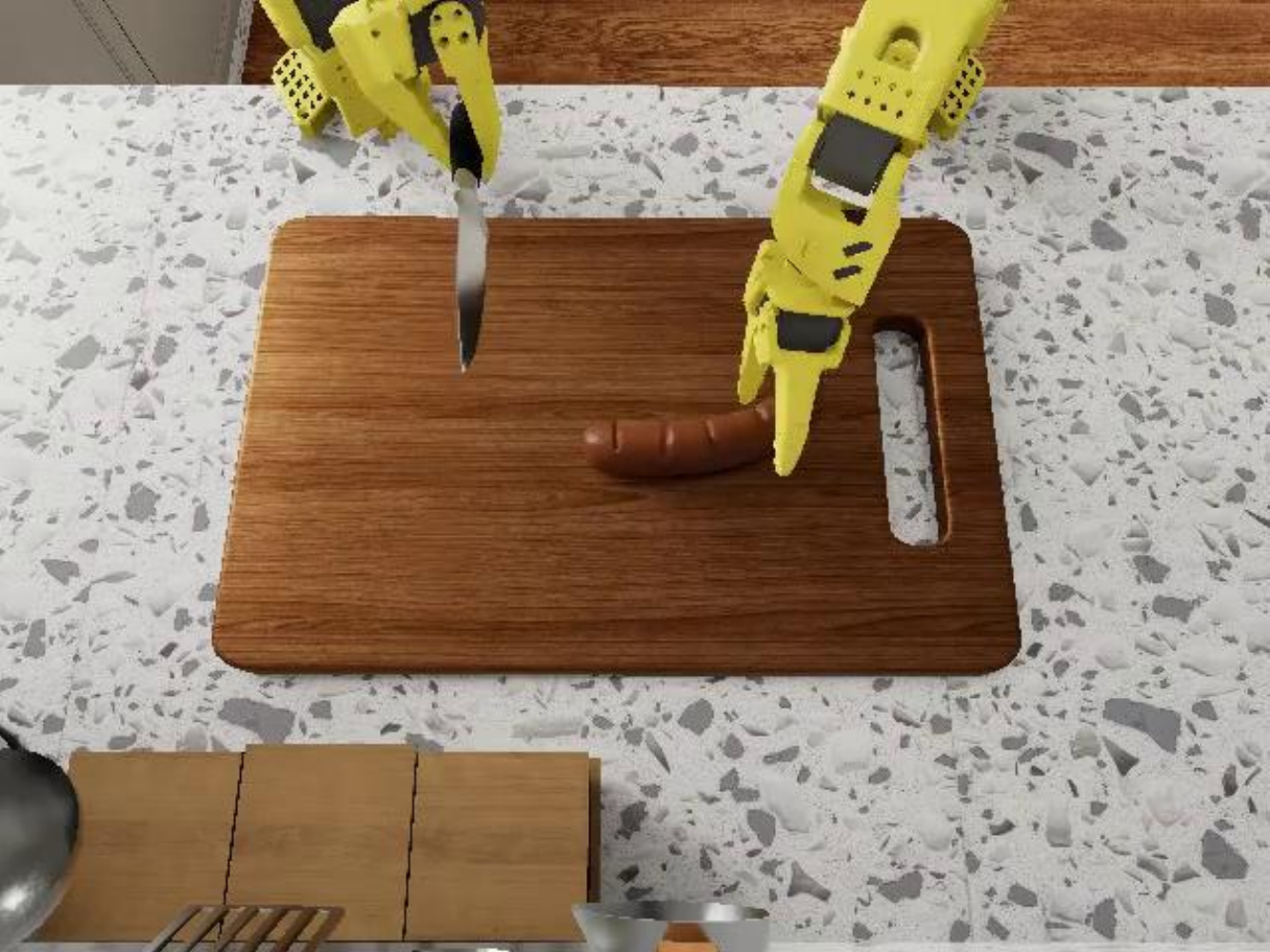}
        \caption{Cut Sausage}
        \label{fig:Sausage}
    \end{subfigure}
    \begin{subfigure}[t]{0.32\linewidth}
        \centering
        \includegraphics[width=\linewidth, keepaspectratio]{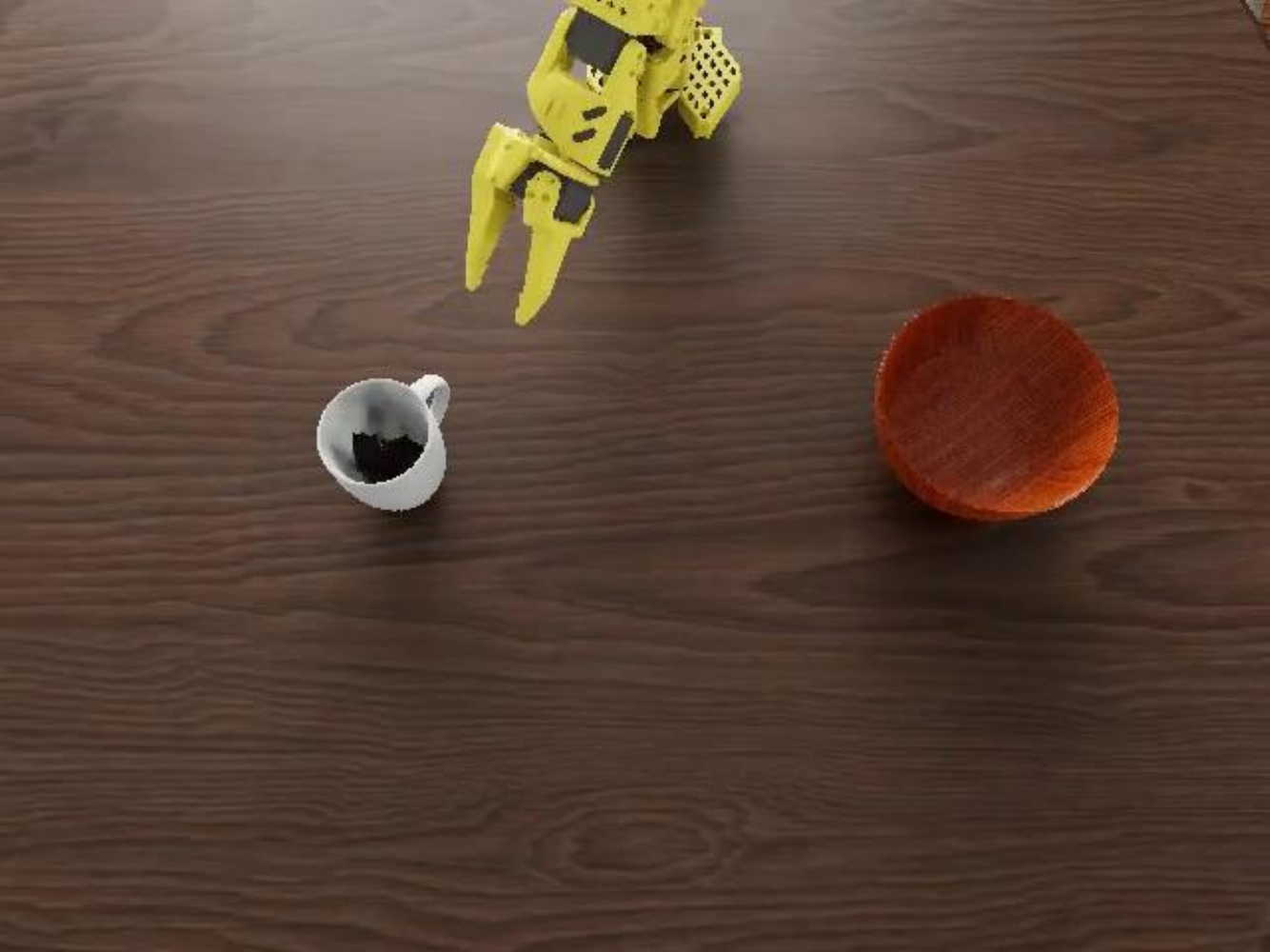}
        \caption{Pour Coffee}
        \label{fig:Pouring}
    \end{subfigure}
    \begin{subfigure}[t]{0.32\linewidth}
        \centering
        \includegraphics[width=\linewidth, keepaspectratio]{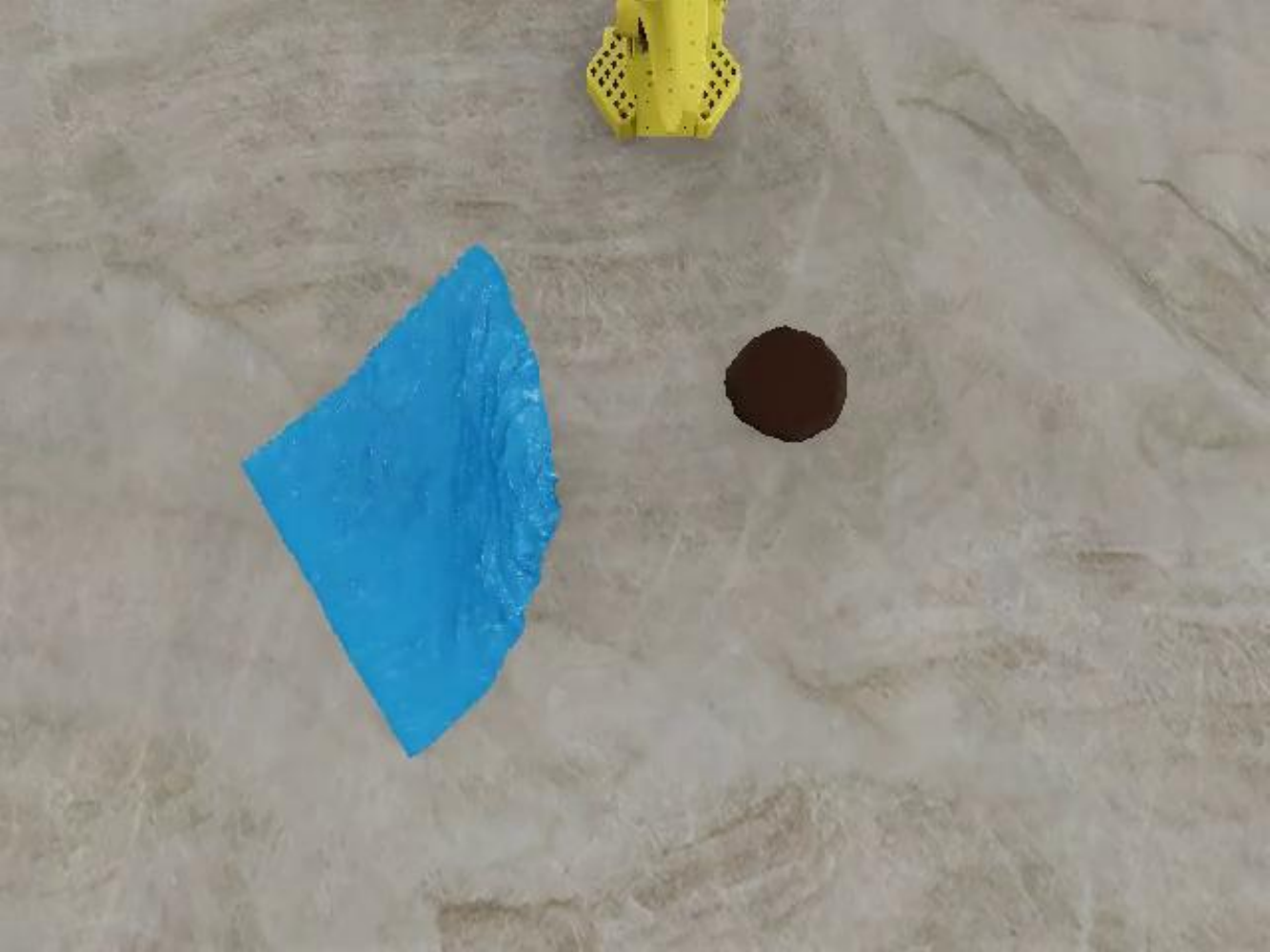}
        \caption{Wipe Surface}
        \label{fig:Wirping}
    \end{subfigure}

    \vspace{-2pt}
    \caption{Gallery of \textbf{Evaluated Tasks}.}
    \label{fig:tasks_visualization}
\end{figure}

\begin{enumerate}
    \item (bedroom) \textbf{Fold Garment}~\ref{fig:fold} requires the dual-arm robot to neatly fold a piece of clothing placed on the table.
    
    \item (bedroom) \textbf{Fling Garment}~\ref{fig:fling} requires the robot to grasp and fling wrinkled clothes to fully unfold and flatten them on the table surface.
    
    \item (kitchen) \textbf{Assemble Burger}~\ref{fig:burger} requires the robot to stack the patty on top of the cheese and bread slices to complete a burger assembly.
    
    \item (kitchen) \textbf{Cut Sausage}~\ref{fig:Sausage} requires the robot to use a knife to cut the sausage into two separate pieces, demonstrating the cutting mechanism in simulation.
    
    \item (living room) \textbf{Pour Coffee}~\ref{fig:Pouring} requires the robot to control the cup orientation to pour coffee into a bowl without spilling.
    
    \item (bathroom) \textbf{Wipe Surface}~\ref{fig:Wirping} requires the robot to use a cloth to wipe water off the floor surface.
\end{enumerate}

\begin{figure}[htbp]
    \centering
    \includegraphics[width=0.45\textwidth]{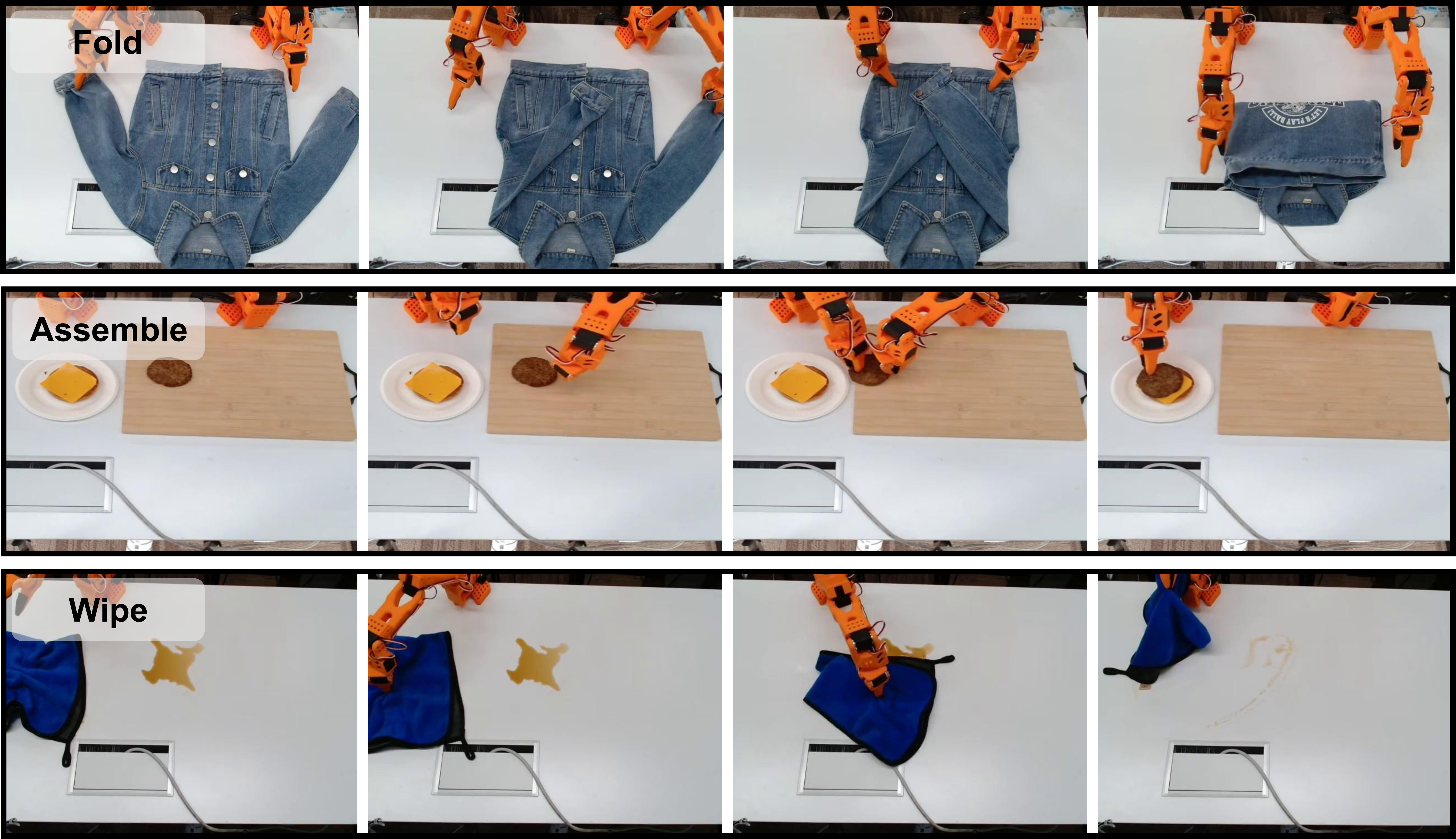}
    \caption{\textbf{Real-world execution on diverse household tasks.}}

    \label{fig:real_exp}
\end{figure}

\subsection{Baselines}

We compare four representative policies under a unified imitation-learning setup.
To analyze the effect of language conditioning, we include two RGB-only imitation baselines, \textbf{Diffusion Policy (DP)}~\cite{chi2023diffusion} and \textbf{ACT}~\cite{zhao2023learning}, and two language-conditioned VLA policies, \textbf{Pi0}~\cite{black2024pi_0} and \textbf{SmolVLA}~\cite{shukor2025smolvla}.

\begin{enumerate}
    \item \textbf{Diffusion Policy (DP)}~\cite{chi2023diffusion}, a generative policy leveraging diffusion models to propose action trajectories.
    \item \textbf{ACT}~\cite{zhao2023learning}, transformer-based policy architecture that predicts sequences of action chunks. 
     \item \textbf{Pi0}~\cite{black2024pi_0}, a flow-matching VLA policy leveraging a pretrained vision-language backbone to generate robot action sequences.
    \item \textbf{SmolVLA}~\cite{shukor2025smolvla}, a lightweight VLA developed by the Hugging Face team.
\end{enumerate}

\subsection{Simulation Experiments} 

Table~\ref{tab:benchmark} reports simulation success rates on six tasks. For fair comparison, all policies are trained with the same data budget (50 teleoperated demonstrations per task) and evaluated over 100 test trials. 

The results show clear task-dependent strengths: SmolVLA performs best on garment manipulation tasks, DP performs best on Cut Sausage, and ACT is competitive on Assemble Burger and Pour Coffee. Notably, Fling Garment is consistently difficult for all methods, with success rates remaining low (ACT: 25.0\%, DP: 20.0\%, SmolVLA: 36.0\%, Pi0: 14.0\%). This indicates that large-deformation, long-horizon garment manipulation is still a major challenge. Overall, these results suggest that LeHome provides a competitive evaluation setting that can effectively distinguish the performance of different policies.
\begin{table}[!t]
    \centering
    \normalsize
    \setlength\tabcolsep{4pt}
    \renewcommand{\arraystretch}{1.2}
    \begin{threeparttable}
    \captionsetup{width=\linewidth}
    \caption{\textbf{Results of Simulation Evaluation.}}
    \label{tab:benchmark}
    \begin{tabular}{lcccc}
    \toprule
    Task & ACT & DP & SmolVLA & Pi0 \\ 
    \midrule
    Fold Garment & 45.0\% & 30.0\% & 70.0\% & 44.0\% \\
    Assemble Burger & 78.0\% & 35.0\% & 40.0\% & 39.0\% \\
    Fling Garment & 25.0\% & 20.0\% & 36.0\% & 14.0\% \\
    Cut Sausage & 77.0\% & 93.0\% & 90.0\% & 75.0\% \\
    Pour Coffee & 80.0\% & 30.0\% & 40.0\% & 40.0\% \\
    Wipe Surface & 60.0\% & 30.0\% & 60.0\% & 60.0\% \\
    \bottomrule
    \end{tabular}
    \end{threeparttable}
\end{table}

\begin{table}[!t]
\centering
\normalsize
\setlength{\tabcolsep}{2pt}
\caption{\textbf{Results of Real-World Experiment.}}
\label{tab:sim2real}
\begin{tabular}{lcccc}
    \toprule
    \multirow{1}{*}{} &
    \multicolumn{2}{c}{ACT} & \multicolumn{2}{c}{SmolVLA} \\
    \cmidrule(lr){2-3} \cmidrule(lr){4-5}
    Task &
     Real & Co-Training & Real & Co-Training \\
    \cmidrule(lr){1-5}
    Fold Garment & 2/10 & \textbf{5/10} & 2/10& \textbf{4/10}\\     
    Assemble Burger &2/10 &\textbf{4/10} & 1/10 &\textbf{4/10} \\   
    Wipe Surface & 1/10 &\textbf{7/10}& 1/10& \textbf{6/10}\\   
    \bottomrule
\end{tabular}
\end{table}

\subsection{Real-World Experiments} 

To evaluate the visual and physical fidelity of LeHome and its contribution to real-world manipulation, we conducted experiments with two LeRobots to assess whether incorporating simulation data can improve policy performance. 

Using dual-arm LeRobot, we compare two training settings: 
(1) \textbf{Real}, where policies are trained using 10 real-world demonstrations only; and 
(2) Sim+Real \textbf{Co-Training}, where policies are trained with both LeHome simulation demonstrations and the same 10 real-world demonstrations. 
We evaluate ACT and SmolVLA under both settings, and report real-world results in Table~\ref{tab:sim2real}, with qualitative examples shown in Fig.~\ref{fig:real_exp}. 
Across all evaluated tasks, \textbf{Sim+Real Co-Training} consistently outperforms \textbf{Real-Only}. On average, the success rate improves from about 15\% to about 50\%, suggesting that LeHome pretraining improves data efficiency and real-world robustness in low-data settings.

\section{CONCLUSIONS}

In this paper, we present LeHome, a comprehensive household simulation environment and benchmark for manipulating deformable objects with everyday items. LeHome provides a diverse suite of physically grounded deformable assets embedded within complete household scenarios, enabling realistic and reproducible experimentation. To expand the application of embodied intelligence to price-sensitive households, we leverage the low-cost LeRobot platform to execute everyday tasks and establish an end-to-end tool chain for training and validating manipulation algorithms.

\bibliographystyle{plain}  
\bibliography{main}

\end{document}